\documentclass[11pt]{article}

\usepackage{arxiv}

\usepackage[utf8]{inputenc}
\usepackage[T1]{fontenc}
\usepackage{float}
\usepackage{url}
\usepackage{booktabs}
\usepackage{amsfonts}
\usepackage{amsmath}
\usepackage{amssymb}
\usepackage{nicefrac}
\usepackage{pifont}
\usepackage{microtype}
\usepackage{xcolor}
\usepackage{graphicx}
\usepackage{subcaption}
\usepackage{multirow}
\usepackage{algorithm}
\usepackage{algorithmic}
\usepackage{pgfplots}
\usepackage{tikz}
\usetikzlibrary{positioning, arrows.meta, shapes.geometric, fit, calc}
\pgfplotsset{compat=1.18}

\newcommand{\cmark}{\ding{51}}
\newcommand{\xmark}{\ding{55}}

\title{BitSkip: An Empirical Analysis of Quantization and Early Exit Composition in Transformers}

\author{
  Ramshankar Bhuvaneswaran \\
  \texttt{bhuvaneshwaran.r@northeastern.edu} \\
  \And
  Handan Liu \\
  \texttt{h.liu@northeastern.edu} \\
}

\date{}

\begin{document}

\maketitle

\begin{abstract}
Ternary weight quantization and early exit mechanisms are different techniques, where one is to compress the model, the other shortens the inference path, yet their interaction when applied jointly remains poorly understood. We present an empirical study of their composition in a causal language model with 1.58-bit (ternary) weights. Using a factorial ablation over the Hadamard transformation and early exit, we train four model variants on WikiText-2 and analyze perplexity, weight distributions, and per-layer exit quality. We found that the techniques do not work well: Hadamard transformation alone improves 1.58 bit quantized model perplexity by 19\% over baseline without hadamard, but adding early exit reverses this progress, making the final model worse than if we had stopped after applying Hadamard. We traced this to gradient interference from the shared language model head and show that intermediate-layer representations are insufficiently mature for accurate next-token prediction under ternary quantization. These results suggest that efficiency techniques should not be assumed to compose favorably and motivate the development of per-layer exit heads for low-bit models.
\end{abstract}
{\noindent\textbf{Code:} \url{https://github.com/Ramshankar07/bitskip}}

\section{Introduction}
The inflation of Large Language Models has reached a point where physical hardware constraints like memory bandwidth, computational throughput, and power consumption. This reality has led researchers to shift focus toward efficiency techniques, with quantization and dynamic neural networks emerging as two prominent approaches.

The weight quantization methods such as AWQ~\cite{lin2024awq} and GPTQ~\cite{frantar2023gptq}, which enable models to run on consumer devices by compressing parameters to low-bit representations. A parallel line of work, early exit~\cite{xin2020deebert, schuster2022calm, elhoushi2024layerskip}, reduces computation by allowing models to terminate inference early when confidence is high. While both approaches have proven effective in isolation, their combined potential remains underexplored. In this work, we investigate whether pushing inference efficiency further—by applying both aggressive low-bit quantization and dynamic early exiting to the same model to see can compound their benefits, or whether their interaction introduces unforeseen trade-offs.

We present \textbf{BitSkip}, an empirical study of how ternary (1.58-bit) weight quantization and early exit compose in a transformer based language model. Our architecture combines BitLinear ternary quantization~\cite{wang2023bitnet}, Hadamard transformation for distribution smoothing~\cite{chee2024quip}, grouped query attention~\cite{ainslie2023gqa}, and a LayerSkip-style~\cite{elhoushi2024layerskip} early exit mechanism with curriculum-based layer dropout.

Our contributions are:
\begin{enumerate}
    \item A \textbf{factorial ablation} over $\{\text{Hadamard}, \text{No Hadamard}\} \times \{\text{Early Exit}, \text{No Early Exit}\}$, isolating the effect of each technique and their interaction.
    \item \textbf{Experiments of the composition}: Hadamard as a standalone achieves the best perplexity (185.46), while the full BitSkip model (Hadamard + early exit) degrades to 216.06, a 17\% performance degradation compared from the Hadamard-only variant.
    \item \textbf{Weight distribution and exit-layer analysis} Shows that Hadamard transformation smoothens which is helpful for ternary weight distributions as mention in ~\cite{chee2024quip} and that early exit representations exhibit a huge difference at the final layer (PPL 12{,}418 at layer 10 vs.\ 216 at layer 11), indicating the shared LM head is ineffective for intermediate predictions in smaller models.
\end{enumerate}

\section{Related Work}

\subsection{Low-Bit Weight Quantization}

Quantization reduces the numerical precision of model parameters to improve efficiency. The BitNet~\cite{wang2023bitnet} from Microsoft introduced ternary weight quantization for transformers, training with weights constrained to $\{-1, 0, +1\}$ via the straight-through estimator (STE)~\cite{bengio2013ste}. BitNet b1.58~\cite{ma2024era} demonstrated that 1.58-bit models can match full-precision performance at scale. Post-training approaches include GPTQ~\cite{frantar2023gptq} and AWQ~\cite{lin2024awq}, which calibrate quantization parameters on a small dataset. QuIP~\cite{chee2024quip} showed that Hadamard incoherence processing smooths weight distributions before quantization, improving accuracy at extreme bit widths. SmoothQuant~\cite{xiao2023smoothquant} addresses activation outliers by migrating quantization difficulty from activations to weights. Our work builds on BitNet-style quantization-aware training\cite{jacob2018quantization}.

\subsection{Early Exit and Adaptive Computation}

The idea of letting a model stop early is skipping deeper layers when an easy input doesn't need them. Graves~\cite{graves2016adaptive} explored adaptive computation time in recurrent networks, and BranchyNet~\cite{teerapittayanon2016branchynet} attached auxiliary classifiers to intermediate layers of CNNs. DeeBERT~\cite{xin2020deebert} brought this to transformers by adding exit classifiers at each BERT layer, while CALM~\cite{schuster2022calm} generalized the approach to sequence-to-sequence models using learned confidence thresholds. More recently, LayerSkip~\cite{elhoushi2024layerskip} showed that training with a curriculum-based layer dropout schedule produces models that naturally tolerate early exit, and pairs this with self-speculative decoding at inference for additional speedup. Depth-adaptive transformers~\cite{elbayad2020depth} take a slightly different angle, learning per-token distributions rather than just relying on a fixed confidence threshold. Our early exit mechanism draws most directly from LayerSkip's curriculum training strategy, where a single shared LM head serves all exit points rather than requiring per-layer classifiers.

\subsection{Composition of Efficiency Techniques}

While individual efficiency techniques are well-studied, their composition hasn't received much attention. Recent surveys~\cite{zhou2024survey, wan2024efficient} catalogue techniques but rarely evaluate interactions. To our knowledge, no prior work systematically studies how ternary quantization and early exit interact in a single model. This gap motivates our factorial ablation study.

\section{Method}

\subsection{Architecture Overview}

The BitSkip model we present is a decoder-only transformer language model. Each of the $L{=}12$ layers consists of a grouped query attention (GQA) block~\cite{ainslie2023gqa} with rotary position embeddings (RoPE)~\cite{su2024roformer}, a feed-forward network (FFN) with squared ReLU activation, sublayer normalization with residual connections, and an optional routing module for early exit decisions. All linear projections use either BitLinear or H-BitLinear layers, described below.

\begin{figure}[t]
\centering
\begin{tikzpicture}[
    node distance=0.6cm and 1.2cm,
    block/.style={rectangle, draw, rounded corners, minimum height=0.8cm, minimum width=2.4cm, align=center, font=\small},
    quantblock/.style={block, fill=blue!10},
    exitblock/.style={block, fill=orange!15},
    arrow/.style={-{Stealth[length=2mm]}, thick},
    label/.style={font=\scriptsize, text=gray}
]
    \node[block] (emb) {Token Embedding\\+ RoPE};

    \node[quantblock, above=of emb] (attn) {GQA Attention\\(BitLinear / H-BitLinear)};
    \node[block, above=of attn] (norm1) {SubLN + Residual};
    \node[quantblock, above=of norm1] (ffn) {FFN\\(BitLinear / H-BitLinear)};
    \node[block, above=of ffn] (norm2) {SubLN + Residual};

    \node[exitblock, right=1.5cm of norm2] (route) {Routing\\Module};
    \node[exitblock, above=of route] (exit) {Early Exit\\(Shared LM Head)};

    \node[block, fill=green!10, left=1.5cm of attn] (skip) {Layer Skip\\$p_l = p_{\max}(l/L)^2$};

    \node[block, above=of norm2] (next) {Next Layer / LM Head};

    \node[draw, dashed, gray, rounded corners, fit=(attn)(norm1)(ffn)(norm2), inner sep=0.3cm, label={[label]above left:$\times L$ layers}] (tblock) {};

    \draw[arrow] (emb) -- (attn);
    \draw[arrow] (attn) -- (norm1);
    \draw[arrow] (norm1) -- (ffn);
    \draw[arrow] (ffn) -- (norm2);
    \draw[arrow] (norm2) -- (next);
    \draw[arrow] (norm2) -- (route);
    \draw[arrow] (route) -- node[right, font=\scriptsize] {$p_{\text{exit}} > 0.5$} (exit);
    \draw[arrow, dashed] (skip) -- (attn);
\end{tikzpicture}
\caption{BitSkip architecture. Each transformer block uses quantized linear layers (BitLinear or H-BitLinear). The routing module produces exit probabilities; during inference, the model exits early if $p_{\text{exit}} > 0.5$. Layer skipping applies curriculum-based dropout during training only.}
\label{fig:architecture}
\end{figure}

\subsection{Ternary Weight Quantization}

\paragraph{BitLinear} Consistent with BitNet~\cite{wang2023bitnet} paper, we quantize weights to ternary values $\{-1, 0, +1\}$. Given a weight matrix $\mathbf{W}$, we compute the scale $\alpha = \text{mean}(|\mathbf{W}|)$ , and quantization is performed as follows:
\begin{equation}
    \mathbf{W}_q = \begin{cases}
        +1 & \text{if } \mathbf{W}_{ij} > 0.5\alpha \\
        -1 & \text{if } \mathbf{W}_{ij} < -0.5\alpha \\
        \phantom{+}0 & \text{otherwise}
    \end{cases}, \qquad
    \hat{\mathbf{W}} = \alpha \cdot \mathbf{W}_q
    \label{eq:ternary}
\end{equation}
Gradients flow through the straight-through estimator: $\hat{\mathbf{W}} = \mathbf{W} - \text{sg}(\mathbf{W}) + \text{sg}(\hat{\mathbf{W}})$, where $\text{sg}(\cdot)$ denotes stop-gradient.

Activations are quantized per-token to $b$-bit integers ($b{=}8$ for BitLinear):
\begin{equation}
    s = \max_{j} |x_j|, \quad
    \mathbf{x}_q = \text{round}\!\left(\frac{(2^{b-1}-1) \cdot \mathbf{x}}{s}\right), \quad
    \hat{\mathbf{x}} = \frac{s \cdot \mathbf{x}_q}{2^{b-1}-1}
    \label{eq:actquant}
\end{equation}

\paragraph{H-BitLinear (Hadamard variant).} Inspired by QuIP~\cite{chee2024quip} and BitNet v2~\cite{wang2025bitnetv2}, we apply a Hadamard transformation before quantization to smooth the activation distribution. BitNet v2 introduced the H-BitLinear module specifically for this purpose in 1-bit LLMs. The forward pass is:
\begin{equation}
    \mathbf{y} = \sigma^2_{\text{ReLU}}\!\Big(\hat{\mathbf{W}}_q \cdot \text{Quant}_{4\text{-bit}}\big(\text{FWHT}(\text{Pad}(\text{LN}(\mathbf{x})))\big)\Big)
    \label{eq:hbitlinear}
\end{equation}
where $\text{FWHT}$ is the Fast Walsh--Hadamard Transform with $O(n \log n)$ complexity, $\text{Pad}$ zero-pads to the next power of two, $\text{LN}$ is layer normalization, and $\sigma^2_{\text{ReLU}}(\mathbf{x}) = \text{ReLU}(\mathbf{x})^2$ is the squared ReLU activation~\cite{so2021primer}.

\subsection{Early Exit with Shared LM Head}

We used approach from LayerSkip~\cite{elhoushi2024layerskip} in using a shared language model head across all exit points. At each layer $l \in \{0, \ldots, L{-}1\}$, the hidden state $\mathbf{h}_l$ is projected through the same LM head to produce logits, and a cross-entropy loss is computed against the target tokens.

The early exit loss which we configured uses early load weighting where earlier layers receive proportionally higher relative weight:
\begin{equation}
    w_l = \frac{1}{l+1}, \quad \bar{w}_l = \frac{w_l}{\sum_{l'=0}^{L-1} w_{l'}}, \quad
    \mathcal{L}_{\text{ee}} = \sum_{l=0}^{L-1} \bar{w}_l \cdot \text{CE}(\text{LMHead}(\mathbf{h}_l),\, \mathbf{y})
    \label{eq:eeloss}
\end{equation}

\subsection{Curriculum-Based Layer Skipping}

During training, we apply per-layer dropout following a quadratic schedule:
\begin{equation}
    p_l = p_{\max} \cdot \left(\frac{l}{L}\right)^2
    \label{eq:layerskip}
\end{equation}
where $p_{\max}$ is the maximum skip probability. Layer $L{-}1$ (the final layer) is never skipped. At each training step, layer $l$ is stochastically skipped with probability $p_l$ independently per sample. Layer skipping is disabled during inference.

\subsection{Routing Module}

Each transformer block includes a learnable gating network that produces exit probabilities:
\begin{equation}
    \text{Gate}_l(\mathbf{x}) = \sigma\!\Big(\mathbf{W}_2 \cdot \text{ReLU}\big(\mathbf{W}_1 \cdot \text{LN}(\mathbf{x})\big)\Big)
    \label{eq:routing}
\end{equation}
where $\mathbf{W}_1 \in \mathbb{R}^{(d/4) \times d}$, $\mathbf{W}_2 \in \mathbb{R}^{1 \times (d/4)}$, and $\sigma$ is the sigmoid function. During training, the exit decision is sampled from a Bernoulli distribution using the straight-through estimator. During inference, the model exits when $p_{\text{exit}} > 0.5$.

\subsection{Training Objective}

The full training loss combines four components:
\begin{equation}
    \mathcal{L} = \mathcal{L}_{\text{task}} + \lambda_{\text{ee}} \cdot \mathcal{L}_{\text{ee}} + \lambda_q \cdot \mathcal{L}_{\text{quant}} + \lambda_r \cdot \mathcal{L}_{\text{route}}
    \label{eq:totalloss}
\end{equation}

\noindent where $\mathcal{L}_{\text{task}}$ is the standard causal LM cross-entropy, $\mathcal{L}_{\text{ee}}$ is the early exit loss (Eq.~\ref{eq:eeloss}), $\mathcal{L}_{\text{quant}}$ is the mean-squared error between full-precision and ternary weights across all quantized layers:
\begin{equation}
    \mathcal{L}_{\text{quant}} = \frac{1}{N} \sum_{i=1}^{N} \|\mathbf{W}_i - \hat{\mathbf{W}}_i\|_2^2
    \label{eq:quantloss}
\end{equation}
and $\mathcal{L}_{\text{route}}$ is a target cost loss encouraging the model to exit at a target layer on average:
\begin{equation}
    p_{\text{exit at } l} = p_l^{\text{exit}} \prod_{i<l}(1 - p_i^{\text{exit}}), \quad
    \bar{l} = \sum_{l=0}^{L-1} l \cdot p_{\text{exit at } l}, \quad
    \mathcal{L}_{\text{route}} = (\bar{l} - l^*)^2
    \label{eq:routeloss}
\end{equation}
where $l^* = L/2$ is the target exit layer.

\section{Experimental Setup}

\subsection{Model Configuration}

Our model loosely follows the Llama 3.1 architecture~\cite{dubey2024llama}, but we made a few deliberate choices for our setting. We use grouped-query attention (GQA)~\cite{ainslie2023gqa} with 8 query heads and 4 key-value heads, which cuts the KV cache in half relative to standard multi-head attention without hurting quality. All core dimensions are made as powers of two so that the fast Walsh-Hadamard transform (FWHT) in H-BitLinear doesn't require any padding, avoiding wasted compute. The full configuration is summarized in Table~\ref{tab:modelconfig}.
\begin{table}[t]
\centering
\caption{Model configuration (85M parameters).}
\label{tab:modelconfig}
\begin{tabular}{lc}
\toprule
\textbf{Parameter} & \textbf{Value} \\
\midrule
Hidden size ($d$) & 512 \\
Layers ($L$) & 12 \\
Attention heads (Q) & 8 \\
KV heads & 4 \\
Head dimension & 64 \\
FFN intermediate & 2048 \\
Vocabulary size & 50{,}257 \\
Max sequence length & 512 \\
Weight bits & 1.58 (ternary) \\
Activation bits & 8 \\
Total parameters & 34{,}603{,}008 (quantizable) \\
\bottomrule
\end{tabular}
\end{table}

\subsection{Training Details}

We train all variants on WikiText-2~\cite{merity2017pointer} with the same setup to keep comparisons fair. The optimizer is AdamW~\cite{loshchilov2019adamw} with a learning rate of $6\!\times\!10^{-4}$, and batch size of 128 with gradient accumulation over 4 steps to simulate a larger effective batch without extra memory pressure. The learning rate follows a warmup-stable-decay (WSD) schedule. We cap runs at 10{,}000 steps and stop early if validation loss hasn't improved for 3 consecutive evaluations. 

\subsection{Ablation Design}                                                            
                                                                                          
  \paragraph{Hyperparameter selection.}                                                   
  We select auxiliary loss weights via a two-stage coordinate sweep on WikiText-2 (500  
  training steps, seed 42).
  In Stage~1, we sweep $\lambda_r \in \{0, 0.01, 0.05, 0.1, 0.2\}$ with $\lambda_q{=}0$,
  selecting $\lambda_r{=}0.2$ (Final Val PPL 251.11).
  In Stage~2, we fix $\lambda_r{=}0.2$ and sweep $\lambda_q \in \{0, 0.01, 0.05, 0.1,
  0.2\}$, selecting $\lambda_q{=}0.05$ (Final Val PPL 254.10).
  A third stage sweeps the maximum skip probability $p_{\max} \in \{0.2, 0.5, 0.7\}$ with
  routing on/off; $p_{\max}{=}0.7$ with routing on yields the best result (Final Val PPL
  252.93).
  The early exit loss weight $\lambda_{\text{ee}}{=}0.3$ and the quadratic dropout
  schedule are adopted from the base BitSkip configuration and held fixed throughout.

  \paragraph{Factorial ablation.}
  Using the selected hyperparameters above, we evaluate the factorial ablation design over two techniques:
  \begin{itemize}
      \item \textbf{Hadamard ablation}: \{No Hadamard (BitLinear), Hadamard (H-BitLinear)\}
      \item \textbf{Early Exit ablation}: \{No Early Exit ($\lambda_{\text{ee}}{=}0$,
  $p_{\max}{=}0$), Early Exit ($\lambda_{\text{ee}}{=}0.3$, $p_{\max}{=}0.7$)\}
  \end{itemize}
  All four configurations share $\lambda_r{=}0.2$, $\lambda_q{=}0.05$, seed 7, and are
  trained for 10k steps with early stopping (patience 3, evaluated every 100 steps) on the
   85M model.
\section{Results}

\subsection{Composition Ablation}

Table~\ref{tab:composition} are the main result for this study. Hadamard Transformation only model achieves the best validation perplexity (185.46) as we know from ~\cite{wang2025bitnetv2}, a 19\% improvement over baseline. Early exit only model is the worst performer (252.37), and the full BitSkip combination (216.06) underperforms the Hadamard-only variant by 17\%.

\begin{table}[t]
\centering
\caption{Composition ablation on WikiText-2.}
\label{tab:composition}
\begin{tabular}{lcccc}
\toprule
\textbf{Configuration} & \textbf{Hadamard} & \textbf{Early Exit} & \textbf{Best Val PPL} & \textbf{Steps} \\
\midrule
Baseline & \xmark & \xmark & 228.77 & 1800 \\
H-only & \cmark & \xmark & \textbf{185.46} & 1800 \\
EE-only & \xmark & \cmark & 252.37 & 2100 \\
BitSkip (H+EE) & \cmark & \cmark & 216.06 & 1800 \\
\bottomrule
\end{tabular}
\end{table}

\begin{figure}[t]
\centering
\begin{tikzpicture}
\begin{axis}[
    ybar,
    bar width=18pt,
    width=0.85\linewidth,
    height=5.5cm,
    ylabel={Validation Perplexity},
    symbolic x coords={Baseline, H-only, EE-only, BitSkip},
    xtick=data,
    ymin=150, ymax=280,
    nodes near coords,
    nodes near coords align={vertical},
    every node near coord/.append style={font=\small},
    enlarge x limits=0.2,
    ylabel style={font=\small},
    xticklabel style={font=\small},
]
\addplot[fill=blue!40] coordinates {
    (Baseline, 228.77)
    (H-only, 185.46)
    (EE-only, 252.37)
    (BitSkip, 216.06)
};
\end{axis}
\end{tikzpicture}
\caption{Validation perplexity across the four composition variants. H-only achieves the best result; adding early exit degrades performance.}
\label{fig:composition_bar}
\end{figure}

\subsection{Full-Precision vs.\ Ternary Degradation}

To understand how each configuration holds up under quantization, This Table~\ref{tab:degradation} compares full-precision and 1.58-bit perplexity for each configuration. The most striking result is BitSkip's degradation: BitSkip suffers the most degradation (+254\%) than any other, while EE-only has the least (+90\%). This suggests that the combination of Hadamard rotation and early exit amplifies quantization sensitivity.

\begin{table}[t]
\centering
\caption{Full-precision (FP) vs.\ 1.58-bit perplexity and degradation.}
\label{tab:degradation}
\begin{tabular}{lccc}
\toprule
\textbf{Configuration} & \textbf{FP PPL} & \textbf{1.58-bit PPL} & \textbf{$\Delta$ PPL (\%)} \\
\midrule
Baseline & 228.76 & 451.03 & +97.2\% \\
H-only & 185.45 & 389.65 & +110.1\% \\
EE-only & 252.34 & 478.38 & +89.6\% \\
BitSkip (H+EE) & 216.06 & 764.36 & +253.8\% \\
\bottomrule
\end{tabular}
\end{table}

\subsection{Weight Distribution Analysis}

Table~\ref{tab:weights} shows how the weight distribution happens across the four model variants. Hadamard transformation raises the mean quantization scale (0.005 $\to$ 0.015), which is consistent with what QuIP ~\cite{chee2024quip} predicts that the transform spreads outlier magnitudes more evenly, giving the ternary quantizer more signal to work with. The EE-only model has the most model shows the most lopsided ternary distribution, while H-only and BitSkip have nearly balanced distributions.

\begin{table}[t]
\centering
\caption{Weight distribution statistics across model variants. ``Hi-Sparse'' counts layers with $\geq$50\% zero weights.}
\label{tab:weights}
\begin{tabular}{lccccc}
\toprule
\textbf{Config} & \textbf{Mean Scale} & \textbf{Sparsity} & \textbf{Hi-Sparse} & \textbf{+1 frac} & \textbf{$-$1 frac} \\
\midrule
Baseline & 0.0051 & 30.4\% & 4 & 34.0\% & 35.6\% \\
H-only & 0.0151 & 35.6\% & 0 & 32.0\% & 32.4\% \\
EE-only & 0.0036 & 26.9\% & 0 & 36.0\% & 37.1\% \\
BitSkip & 0.0122 & 35.3\% & 0 & 32.3\% & 32.4\% \\
\bottomrule
\end{tabular}
\end{table}

\subsection{Per-Layer Exit Perplexity}

This Table~\ref{tab:exitppl} shows the per-layer exit perplexity for the BitSkip (H+EE) model, This is depicts the quadratic schedule from LayerSkip~\cite{elhoushi2024layerskip}. The exit layer perplexity is not a gradual decline; it is a cliff between layer 10 (PPL 12{,}418) and layer 11 (PPL 216), indicating that the shared LM head produces meaningful predictions only at the final layer. Intermediate representations are not sufficiently refined for accurate next-token prediction under ternary quantization.

\begin{table}[t]
\centering
\caption{Per-layer exit perplexity for BitSkip (H+EE). The shared LM head is effective only at the final layer.}
\label{tab:exitppl}
\begin{tabular}{cccccc}
\toprule
\textbf{Layer} & \textbf{Loss} & \textbf{PPL} & \textbf{Layer} & \textbf{Loss} & \textbf{PPL} \\
\midrule
0 & 11.27 & 78{,}480 & 6 & 10.59 & 39{,}540 \\
1 & 11.37 & 86{,}584 & 7 & 10.15 & 25{,}603 \\
2 & 11.33 & 82{,}870 & 8 & 9.92 & 20{,}237 \\
3 & 11.25 & 76{,}889 & 9 & 9.68 & 15{,}923 \\
4 & 11.13 & 68{,}198 & 10 & 9.43 & 12{,}418 \\
5 & 10.97 & 58{,}027 & 11 & 5.38 & \textbf{216} \\
\bottomrule
\end{tabular}
\end{table}

\begin{figure}[t]
\centering
\begin{tikzpicture}
\begin{axis}[
    width=0.85\linewidth,
    height=5.5cm,
    xlabel={Exit Layer},
    ylabel={Validation Perplexity},
    xtick={0,1,...,11},
    ymode=log,
    ymin=100, ymax=100000,
    grid=major,
    grid style={dashed, gray!30},
    mark size=2.5pt,
    xlabel style={font=\small},
    ylabel style={font=\small},
    xticklabel style={font=\small},
    yticklabel style={font=\small},
    legend pos=north east,
    legend style={font=\small},
]
\addplot[color=red, mark=*, thick] coordinates {
    (0, 78480) (1, 86584) (2, 82870) (3, 76889) (4, 68198)
    (5, 58027) (6, 39540) (7, 25603) (8, 20237) (9, 15923)
    (10, 12418) (11, 216)
};
\addlegendentry{BitSkip (H+EE)}
\end{axis}
\end{tikzpicture}
\caption{Per-layer exit perplexity (log scale). A dramatic cliff between layer 10 and 11 shows the shared LM head cannot produce useful predictions from intermediate representations.}
\label{fig:exit_curve}
\end{figure}

\subsection{Hyperparameter Sweeps}
We swept auxiliary loss weights and routing configurations in short 500-step runs to select hyperparameters before committing to the full 10k-step ablation. The details are reported in Tables~\ref{tab:lambda_r}--\ref{tab:routing} in the appendix;
\paragraph{Routing loss ($\lambda_r$).} Table~\ref{tab:lambda_r} shows that $\lambda_r{=}0.2$ achieves the best perplexity (251.11) for the non-Hadamard model. The effect is modest---the range across $\lambda_r \in \{0, 0.01, 0.05, 0.1, 0.2\}$ spans only 254--259 in final validation PPL, suggesting the model is robust to this hyperparameter at 500 training steps.

\begin{table}[H]
\centering
\caption{Routing loss weight ($\lambda_r$) sweep. Non-Hadamard model, 500 steps.}
\label{tab:lambda_r}
\begin{tabular}{ccc}
\toprule
$\lambda_r$ & Best Val PPL & Final Val PPL \\
\midrule
0.0 & 287.34 & 254.56 \\
0.01 & 288.64 & 254.27 \\
0.05 & 289.35 & 259.17 \\
0.1 & 288.24 & 257.29 \\
0.2 & \textbf{283.48} & \textbf{251.11} \\
\bottomrule
\end{tabular}
\end{table}

\paragraph{Quantization loss ($\lambda_q$).} Table~\ref{tab:lambda_q} shows results for the quantization loss sweep. The best value is $\lambda_q{=}0.05$ (final PPL 254.10), though the differences are small.

\begin{table}[H]
\centering
\caption{Quantization loss weight ($\lambda_q$) sweep. Non-Hadamard model, $\lambda_r{=}0.2$, 500 steps.}
\label{tab:lambda_q}
\begin{tabular}{ccc}
\toprule
$\lambda_q$ & Best Val PPL & Final Val PPL \\
\midrule
0.0 & 289.35 & 259.17 \\
0.01 & 289.48 & 255.01 \\
0.05 & \textbf{284.24} & \textbf{254.10} \\
0.1 & 285.95 & 254.29 \\
0.2 & 291.57 & 257.89 \\
\bottomrule
\end{tabular}
\end{table}

\paragraph{Routing configuration.} Table~\ref{tab:routing} compares learned routing against fixed dropout across three skip probabilities. The best result at 500 steps is actually fixed dropout with $p_{\max}{=}0.7$ (PPL 251.43), slightly outperforming learned routing at the same $p_{\max}$ (252.93). Higher skip probability ($p_{\max}{=}0.7$) consistently outperforms lower values.

\begin{table}[H]
\centering
\caption{Routing configuration sweep. Non-Hadamard model, 500 steps.}
\label{tab:routing}
\begin{tabular}{llcc}
\toprule
\textbf{Routing} & $p_{\max}$ & \textbf{Best Val PPL} & \textbf{Final Val PPL} \\
\midrule
OFF & 0.7 & \textbf{284.39} & \textbf{251.43} \\
OFF & 0.5 & 287.34 & 254.56 \\
OFF & 0.2 & 293.30 & 259.85 \\
ON & 0.7 & 286.06 & 252.93 \\
ON & 0.5 & 285.50 & 254.80 \\
ON & 0.2 & 290.34 & 253.43 \\
\bottomrule
\end{tabular}
\end{table}

\begin{table}[H]
\centering
\caption{H-BitLinear collapse diagnosis. Multi-seed golden config, all collapsed before the squared ReLU fix.}
\label{tab:hcollapse}
\begin{tabular}{lccc}
\toprule
\textbf{Experiment} & \textbf{Seed} & \textbf{Best Val PPL} & \textbf{Status} \\
\midrule
Before fix (seed 42) & 42 & 1{,}794.94 & Collapsed \\
Before fix (seed 123) & 123 & 1{,}796.97 & Collapsed \\
Before fix (seed 456) & 456 & 1{,}803.20 & Collapsed \\
\midrule
After fix (seed 42) & 42 & \textbf{957.03} & Learning \\
\bottomrule
\end{tabular}
\end{table}

\section{Discussion}

\paragraph{Why is the composition sub-additive?} We think two things are going on. The first is gradient interference through the shared LM head. Because the early exit loss pushes gradients from all 12 layers into a single set of head parameters, and only the final layer's representations are actually mature enough for next-token prediction Table~\ref{tab:exitppl}, the gradients from earlier layers amount to noise.  Second, \textbf{layer skipping during training} means that intermediate representations are computed less frequently, reducing the number of gradient updates they receive. Combined with ternary quantization noise, this makes intermediate representations even less suitable for early exit.

\paragraph{Practical recommendations.} Based on our findings, we suggest to avoid pairing early exit with a shared LM head at small model scales. In the exit layer's PPL, we observe suggests that a single head simply cannot serve double duty as both a final-layer classifier and an intermediate-layer exit point; per-layer heads or distillation-based exits are likely necessary to make early exit viable in this setting.

\paragraph{Limitations.}
This study has several limitations. (1)~We evaluate at a small model scale (85M parameters) where current models are larger and may exhibit different composition dynamics as representations mature faster across layers. (2)~All experiments use WikiText-2, a relatively small dataset; results may differ on larger dataset (3)~Training was limited to 10{,}000 steps with early stopping; longer training may allow early exit to converge. (4)~We use a shared LM head for early exit rather than per-layer classifiers, which may disadvantage the early exit variants.

\paragraph{On scaling.} We suspect that much of the negative interaction we observe is a consequence of scale or rather, the lack of it. At 85M parameters, the model is operating in a regime where representations across layers are still relatively immature. The exit layer PPL in Table~\ref{tab:exitppl} may say less about early exit as a technique and more about the fact that a small model trained on limited data simply hasn't had enough capacity or gradient signal to develop meaningful intermediate representations. But in larger models trained on substantially more data, layers tend to specialize earlier and hidden states become interpretable at shallower depths which is exactly the condition that makes early exit viable. LayerSkip~\cite{elhoushi2024layerskip} demonstrates this at 7B and 13B scale with full-precision weights. It is plausible that at, say, 1B+ parameters with a larger corpus, the Hadamard and early exit benefits would compound rather than interfere the Hadamard transform would still smooth quantization, while the richer intermediate representations would give the shared head (or per-layer heads) something meaningful to work with. We leave this investigation to future work, but view it as the most promising direction for making BitSkip practical.
\section{Conclusion}

We presented BitSkip, a systematic empirical study of how ternary weight quantization and early exit compose in transformer based language models. Through a $2\!\times\!2$ factorial ablation, we found that:

\begin{itemize}
    \item \textbf{Hadamard Transformation is the strongest technique}, improving perplexity by 19\% (228.77 $\to$ 185.46) by smoothing ternary weight distributions.
    \item \textbf{Early exit hurts when composed with ternary quantization}, degrading the Hadamard-only variant from 185.46 to 216.06 PPL.
    \item \textbf{The composition is sub-additive}: the shared LM head cannot produce useful predictions from intermediate layers under ternary quantization, as evidenced by a perplexity cliff between layer 10 (12{,}418) and layer 11 (216).
\end{itemize}

\bibliographystyle{plain}
\bibliography{references}

\appendix

\section{Architecture Parameter Breakdown}
\label{app:params}

\begin{table}[h]
\centering
\caption{Parameter count breakdown for the 85M model.}
\label{tab:params}
\begin{tabular}{lcc}
\toprule
\textbf{Component} & \textbf{Parameters} & \textbf{Quantizable} \\
\midrule
Token embedding & $50{,}257 \times 512 = 25.7$M & No \\
Attention Q proj ($\times 12$) & $512 \times 512 \times 12 = 3.1$M & Yes \\
Attention K proj ($\times 12$) & $512 \times 256 \times 12 = 1.6$M & Yes \\
Attention V proj ($\times 12$) & $512 \times 256 \times 12 = 1.6$M & Yes \\
Attention O proj ($\times 12$) & $512 \times 512 \times 12 = 3.1$M & Yes \\
FFN up proj ($\times 12$) & $512 \times 2048 \times 12 = 12.6$M & Yes \\
FFN down proj ($\times 12$) & $2048 \times 512 \times 12 = 12.6$M & Yes \\
LM head & $512 \times 50{,}257 = 25.7$M & No \\
LayerNorm, routing, biases & $\sim$0.2M & No \\
\midrule
\textbf{Total} & $\sim$85M & 34.6M \\
\bottomrule
\end{tabular}
\end{table}

\end{document}